\PassOptionsToPackage{table}{xcolor}
\documentclass[10pt,twocolumn,letterpaper]{article}

\usepackage{iccv}              

\usepackage{booktabs}
\usepackage{graphicx}
\usepackage{multirow}  
\usepackage{makecell}
\usepackage{etoc}
\usepackage{titletoc}

\definecolor{iccvblue}{rgb}{0.21,0.49,0.74}
\usepackage[pagebackref,breaklinks,colorlinks,allcolors=iccvblue]{hyperref}

\def\paperID{1433} 
\def\confName{ICCV}
\def\confYear{2025}

\title{\textit{SpatialCrafter}: Unleashing the Imagination of Video Diffusion Models \\ for Scene Reconstruction from Limited Observations}

\author{Songchun Zhang$^{1,2}$ \quad Huiyao Xu$^1$ \quad Sitong Guo$^1$ \quad Zhongwei Xie$^2$ \\ Hujun Bao$^1$ \quad Weiwei Xu$^1$ \quad Changqing Zou$^{1,3}$\thanks{Corresponding author} \\
$^1$ZJU \quad $^2$HKUST \quad $^3$Zhejiang Lab
}

\begin{document}
\twocolumn[{%
    \renewcommand\twocolumn[1][]{#1}%
    \maketitle
    \centering
    \vspace{-0.6cm}

\includegraphics[width=1.0\textwidth]{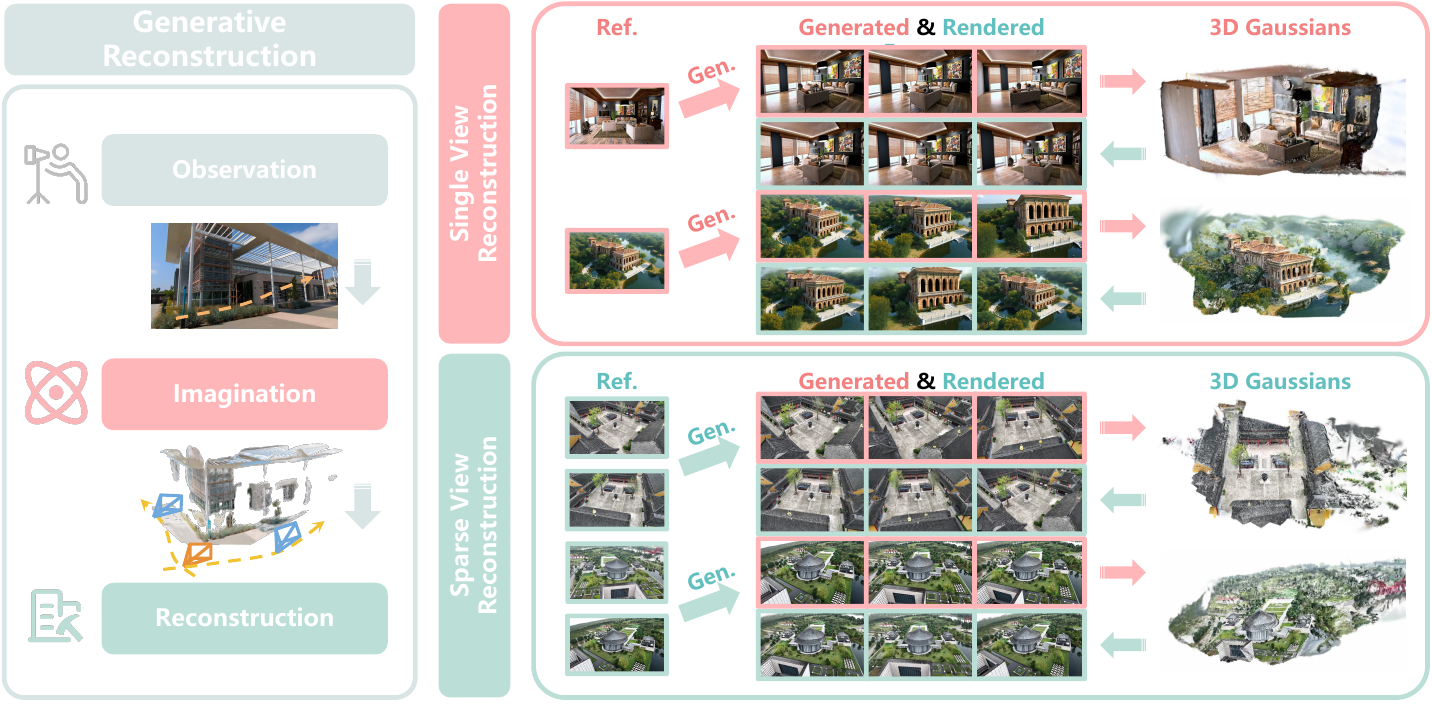}
\captionof{figure}{\textbf{Overview.} Our method, \textit{SpatialCrafter}, generates additional novel views of a scene from few inputs by leveraging camera trajectory-guided video diffusion models, which alleviates the ambiguity of sparse view reconstruction and robustly reconstructs the scene from the generated video frames. It demonstrates impressive performance in both indoor and outdoor scenes, while also exhibiting generalization capabilities in real scenes and synthetic images. More visualization results can be found in the supplementary material.}
    \vspace{1.0cm}
}]

{
\renewcommand{\thefootnote}{}%
\footnotetext{* Corresponding author}%
}

\begin{abstract}
Novel view synthesis (NVS) boosts immersive experiences in computer vision and graphics. Existing techniques, though progressed, rely on dense multi-view observations, restricting their application. 
We tackle the task of reconstructing photorealistic 3D scenes from only one or a few input views.
We introduce SpatialCrafter, a framework that leverages the rich knowledge in video diffusion models to generate plausible additional observations, thereby alleviating reconstruction ambiguity. 
Through a trainable camera encoder and an epipolar attention mechanism for explicit geometric constraints, we achieve precise camera control and 3D consistency, further reinforced by a unified scale estimation strategy to handle scale discrepancies across datasets.
Furthermore, by integrating monocular depth priors with semantic features in the video latent space, our framework directly regresses 3D Gaussian primitives and efficiently processes long-sequence features using a hybrid network structure. 
Extensive experiments show our method enhances sparse view reconstruction and restores the realistic appearance of 3D scenes. Project page: \url{https://franklinz233.github.io/projects/spatialcrafter/}.

\end{abstract}    
\section{Introduction}
\label{sec:intro}


Novel View Synthesis (NVS), as a key technology in computer vision and graphics, provides crucial support for immersive experiences in fields such as video games and mixed reality. 
Although neural reconstruction techniques have made significant progress in recent years~\cite{kerbl20233dgs,wang2021nerf,xiangli2021citynerf,reiser2021kilonerf,muller2022instant}, these methods typically rely on dense multi-view observation data, which face numerous limitations in practical applications.
Therefore, this paper focuses on a more practically valuable challenge: how to achieve high-quality 3D scene reconstruction and synthesize realistic novel views from sparse or even single-view observation.

This problem is hard because (1) real-world scenes are diverse and complex, (2) occluded regions are unseen in sparse captures, and (3) geometric cues are too few to constrain reconstruction.
%
To address these challenges, some methods~\cite{li2024dngaussian,zhu2023fsgs,kim2022infonerf,niemeyer2022regnerf,wang2023sparsenerf,zhang2022nerfusion} constrain the optimization process by introducing regularization terms, but these methods need to be optimized for each scene, are not generalizable, and are difficult to cope with complex scenarios.
%
Recently, some feed-forward methods~\cite{charatan2024pixelsplat,chen2024mvsplat,wang2024dust3r,smart2024splatt3r,hong2024coponerf,szymanowicz24splatterimage,gslrm, lgm} have achieved generalizable sparse view scene reconstruction by directly predicting 3D Gaussian primitives for each pixel. 
However, such methods produce severe artifacts when reconstructing occluded regions and applying them to the extrapolation setting.
%
To address this challenge, some methods~\cite{liu2023zero123,wang2023motionctrl,zeronvs,wu2024reconfusion} have introduced the priors from the diffusion model to achieve sparse or single-view NVS, but they perform poorly on scene-level data and lack precise pose control and 3D consistency.
This is due to the fact that low-dimensional camera information such as Euler angles and extrinsic struggle to provide comprehensive control signals to the generative models.

In this paper, we propose a framework named \textit{SpatialCrafter} for scene reconstruction and novel view synthesis based on sparse or single-view inputs.
%
%
Our innovation lies in leveraging the rich physical world knowledge embedded in video diffusion models to provide plausible additional observations for scene reconstruction, thereby effectively reducing the problem complexity.
%
Specifically, we first focus on enhancing precise camera control and 3D consistency in generated videos.
To achieve this, we parameterize camera settings using ray embeddings~\cite{zhang2024raydiffusion} or metric depth-warped images~\cite{yu2024viewcrafter} and incorporate them into the video diffusion model via a trainable camera encoder.
Furthermore, we propose an epipolar attention mechanism that improves 3D consistency between video frames through explicit geometric constraints.
Previous methods~\cite{wang2023motionctrl, he2024cameractrl} often experience performance degradation and difficulty generating large-motion videos when trained on multiple scene datasets.
We identified that these issues primarily stem from scale ambiguity in scene datasets.
To overcome this, we introduce a unified scale estimator to calibrate scene dataset, enabling effective joint training across multiple datasets.

Although we can generate visually coherent video sequences, relying solely on generated frames to reconstruct general scenes often leads to suboptimal solutions, particularly in large-scale outdoor environments and stylized scenes. 
To address this, we propose incorporating monocular depth priors with rich semantic features extracted from the video latent space. 
Leveraging these latent features, our method directly regresses 3D Gaussian primitives of the scene via a feed-forward network. 
Furthermore, to efficiently handle long-sequence feature interactions, we design a hybrid architecture that integrates Mamba blocks with Transformer blocks.
%
%
%
Experiments show that our method not only improves the sparse-view reconstruction, but also accurately restores the appearance of 3D scenes, especially when extrapolating from a single view and when there are few sparse view overlaps.
In conclusion, our key contributions can be summarized as follows:
\begin{itemize}
    \item
    We introduce a framework that effectively utilizes the physical-world knowledge embedded in video diffusion models to provide additional plausible observations for sparse-view scene reconstruction, thus reducing the ambiguity of sparse view scene reconstruction.
    
    \item To address the scale ambiguity problem that occurs in joint training across datasets, we develop a unified scale estimation approach for trajectory calibration. This solves the performance degradation problem, thus enabling effective multi-dataset training.

    \item We combine monocular depth priors with semantic features extracted from the video latent space, and directly regress 3D Gaussian primitives through a feed-forward manner. Meanwhile, we propose a hybrid architecture integrating Mamba blocks with Transformer blocks to efficiently handle long-sequence feature interactions.
\end{itemize}


\section{Related Work}
\label{sec:related work}

\begin{figure*}[t]
    \centering
    \includegraphics[width=1.0\textwidth]{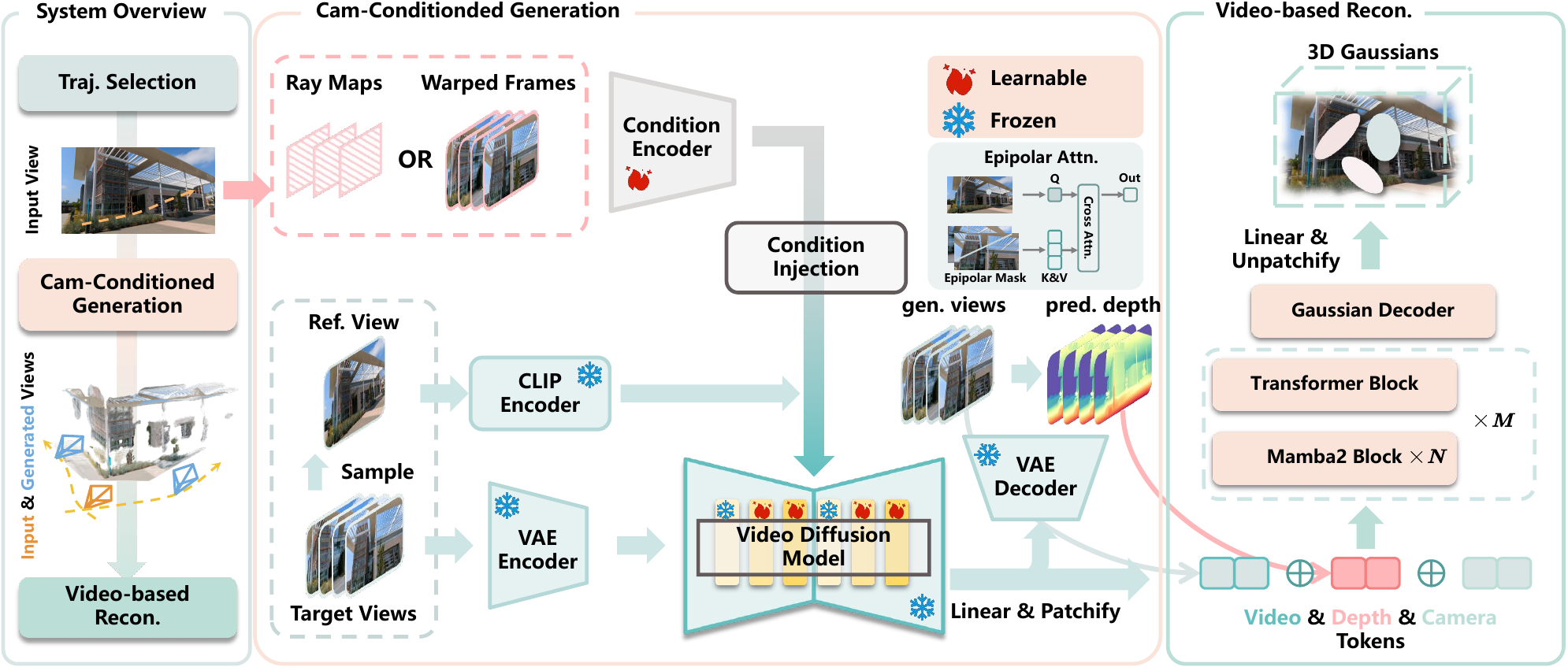}
    \vspace{-2mm}
    \caption{\textbf{Overview of our pipeline.} Our generative reconstruction pipeline consists of two parts: camera-controlled video generation, and then video-based reconstruction. 
    First, we set the exploration path based on the input views. In the video generation module, ray embeddings are used to parameterize the camera trajectories, and a cross-attention mechanism with epipolar constraints is introduced to improve the 3D consistency of the generated video.
    The video-based scene reconstruction pipeline integrates monocular depth priors with semantic features extracted from the video latent space, and directly regress 3D Gaussian primitives via a feed-forward manner.}
    \vspace{-2mm}
    \label{fig:pipeline}
\end{figure*}

\subsection{Sparse-View Scene Reconstruction}
NeRF~\cite{wang2021nerf} and Gaussian Splatting ~\cite{kerbl20233dgs} have achieved photorealistic representations of 3D scenes, but require optimization on densely collected image sets for each scene.
To address this problem, some approaches~\cite{li2024dngaussian,wang2023sparsenerf,zhu2023fsgs,yang2023freenerf} focus on constraining the optimization process of 3D representations via manually designed regularization.
Some other methods~\cite{honglrm,charatan2024pixelsplat,chen2024mvsplat,szymanowicz2024flash3d,xu2024depthsplat,ye2024noposplat, wang2025vggt, zhang2025flare} are trained on large-scale datasets and directly predict the 3D representation in a feed-forward manner.
Recently, some approaches~\cite{fan2024instantsplat, smart2024splatt3r} based on end-to-end 3D reconstruction models~\cite{wang2024dust3r, leroy2024groundingmast3r, charatan2024pixelsplat}, have achieved efficient and high-quality 3D reconstruction.
However, these methods cannot handle occluded regions, and reconstruction failures (e.g., under large viewpoint changes) can severely affect the quality of novel view synthesis.
More relevant to our work, some works~\cite{wang2025videoscene, liu2024reconx, liang2025wonderland} have employed priors from video diffusion models to enhance the performance of sparse-view reconstruction.

\subsection{Diffusion-based Novel View Synthesis}
Since text-to-image diffusion models~\cite{rombach2022high} contain rich natural image priors, some methods~\cite{poole2022dreamfusion, wang2024prolificdreamer} directly optimize 3D representations~\cite{wang2021nerf,kerbl20233dgs} via score distillation sampling~\cite{poole2022dreamfusion}.
Several approaches~\cite{liu2023zero123, zeronvs, tung2024megascenes, chan2023genvs} enable zero-shot novel view synthesis by fine-tuning the image and pose-conditioned generative models on large-scale 3D object~\cite{deitke2023objaverse} or scene~\cite{zhou2018real10k,tung2024megascenes} datasets.
However, they still have difficulty synthesizing consistent novel views.
To solve this, some approaches~\cite{liu2023syncdreamer,zheng2024free3d,gao2024cat3d,muller2024multidiff, shi2023mvdream,wu2024reconfusion,cao2024mvgenmaster} generate multi-views simultaneously and model the correlation between multiple views with ground-truth poses.
Some recent methods~\cite{ma2024see3d,li2025nvcomposer} proposed to use large-scale unposed images to train NVS methods.
In addition, some methods~\cite{chung2023luciddreamer, zhang20243dscenedreamer, fridman2024scenescape, yu2024wonderjourney,shriram2024realmdreamer, engstler2024invisible} utilize depth-based warping to synthesize novel views and employ the T2I model to inpaint the warped images.
However, the novel views generated by these methods tend to suffer from artifacts and cumulative errors, limiting their use in generalized scenes.

\subsection{Controllable Video Generation}
While recent text-to-video diffusion models have achieved remarkable progress, they inherit the controllability limitations of their text-to-image counterparts, often requiring additional conditioning mechanisms to align generated videos with user intent. 
Several works~\cite{wang2024videocomposer,guo2023animatediff,hu2024animateanyone,xu2024magicanimate} have been carried out to introduce a variety of conditions into video generation models.
Recently, attention has focused on controlling the camera motion of the video.
Some approaches~\cite{guo2023animatediff,sun2024dimensionx} employs motion-specific training for predefined camera movements, though this framework struggles with complex trajectory synthesis. 
Some methods~\cite{wang2023motionctrl, he2024cameractrl, he2025cameractrl, xu2024camco, bahmani2024ac3d} enable more complex camera control by parameterizing the camera trajectory and injecting it into a pre-trained video diffusion model via a trainable camera encoder.
Another line of work~\cite{yu2024viewcrafter, liu2024reconx, you2024nvs, ren2025gen3c} advances this direction by enabling pixel-accurate view compositing through a point-based rendering approach.




\section{Method}




This section begins by explaining our camera-controlled video generation model (Section~\ref{sec:video_gen}), which employs ray embeddings for precise pose control.
Then, we describe how to use epipolar geometry constraints to enhance 3D consistency in the video frames.
Next, we introduce ourscene reconstruction pipeline (Section~\ref{sec:scene_recon}), which aims to perform stable 3D reconstruction based on the generated video latents.

\subsection{Camera-Conditioned Video Generation}
\label{sec:video_gen}

\noindent{\textbf{Scale Alignment.}}
To address this challenge, our approach first employs VGGT~\cite{wang2025vggt} to estimate initial camera parameters, represented by the extrinsic matrix $\mathbf{P} = [\mathbf{R} | \mathbf{T}]$, and a corresponding depth map $d_v$ for each video frame. 
To establish a consistent scale across these datasets, we leverage the Metric3D~\cite{yin2023metric3d} to infer a canonical metric depth map, denoted $d_{m}$. 
We then align our initial depth prediction, $d_{v}$, to the canonical scale by calculating a scale factor $s$, defined as the ratio of the inter-percentile ranges (IPR) of the two depth maps:
$ s = \operatorname{IPR}_{0.8, 0.2}(d_{v})/\operatorname{IPR}_{0.8, 0.2}(d_{m})$.
This scale factor is then used to adjust the translation component of the camera extrinsics, resulting in $\mathbf{P}_{scaled} = [\mathbf{R} | s \cdot \mathbf{T}]$. 
This procedure yields camera extrinsics with a consistent absolute scale, effectively resolving alignment issues when fusing data from different sources.

\noindent{\textbf{Camera Injection.}}
Inspired by recent methods~\cite{he2024cameractrl, zhang2024raydiffusion, sitzmann2021lightfield}, we chose to use ray embeddings~\cite{he2024cameractrl} or the depth warping frames~\cite{yu2024viewcrafter} to represent the camera information.
Specifically, for a ray defined by the origin $o\in \mathbb{R}^3$ and the normalized direction $d\in \mathbb{R}^3$, we can represent it as $(o \times d, d) \in \mathbb{R}^6$.
Given the camera parameters, the direction of the ray $d'$ corresponding to the pixel coordinates $(u,v)$ can be calculated as $d' = \mathbf{R} \mathbf{K}^{-1} (u,v,1)^\top + \mathbf{T}$.
For depth warping frames, we leverage the pretrained depth estimation model~\cite{wang2024moge} to map the reference image into a 3D point cloud. Subsequently, we render novel views at specified camera poses by projecting this point cloud via the target camera parameters. 
Afterwards, we use a trainable conditional encoder to inject the camera information into the video model instead of fine-tuning all model parameters.
The training objective is:
\begin{equation}
    \mathcal{L} = \mathbb{E}_{z, z_0, \epsilon, {C}, t} \left[\left\|\epsilon - \epsilon_\theta(z_t; z_0, t, \phi ({C}))\right\|_2^2\right]
\end{equation}
where $\phi ({C})$ represents the camera conditional encodings, $z_0$ denotes the latent embeddings of the reference frame, and $t$ indicates the timestamps.

\noindent{\textbf{Epipolar Feature Aggregation.}}
Camera embedding enhances control but 3D video consistency remains challenging due to dense self-attention allowing unrestricted cross-frame pixel interactions. 
We address this by incorporating epipolar geometric constraints, and aggregating features along epipolar lines.
For a pixel $\boldsymbol{p} = (u, v)$ in frame $i$, its corresponding epipolar line $\boldsymbol{l}_{ik}$ in frame $k$ is computed as $\boldsymbol{l}_{ik}(\boldsymbol{p}) = \boldsymbol{F}_{ik} \cdot \tilde{\boldsymbol{p}}$, where $\tilde{\boldsymbol{p}} = (u, v, 1)^\top$ are the homogeneous coordinates, and $\boldsymbol{F}_{ik} \in \mathbb{R}^{3 \times 3}$ is the fundamental matrix. The fundamental matrix is decomposed as $\boldsymbol{F}_{ik} = \boldsymbol{K}_k^{-\top} \boldsymbol{E}_{ik} \boldsymbol{K}_i^{-1}$, where $\boldsymbol{K}_i, \boldsymbol{K}_k \in \mathbb{R}^{3 \times 3}$ are the camera intrinsics, and $\boldsymbol{E}_{ik}$ is the essential matrix.
We then convert epipolar lines into attention masks by computing per-pixel distances and applying a threshold, restricting attention to geometrically valid regions.

\noindent{\textbf{Sparse-View Setting.}}
To adapt to the sparse view input setting, we formulate the task as a video frame interpolation problem conditioned on the given start and end frames. 
To maximally preserve the priors from the pre-trained models~\cite{blattmann2023svd}, we inject boundary frame conditions in both latent and semantic space.
Specifically, we combine latent features of the first and last frames with their noisy latent representations, and then concatenate the extracted CLIP embeddings from both frames for cross-attentional feature injection.
The training objective is:
\begin{equation}
    \mathcal{L} = \mathbb{E}_{z, z_0, z_n \epsilon, {C}, t} \left[\left\|\epsilon - \epsilon_\theta(z_t; z_0, z_n, t, \phi ({C})))\right\|_2^2\right]
\end{equation}
where $z_n$ and $z_0$ are the latent embeddings of the final and initial frames, respectively.
During the inference phase, we use VGGT~\cite{wang2025vggt} to estimate the extrinsic and intrinsic parameters for the input views.

\begin{figure*}[t]
    \centering
    \includegraphics[width=1.0\textwidth]{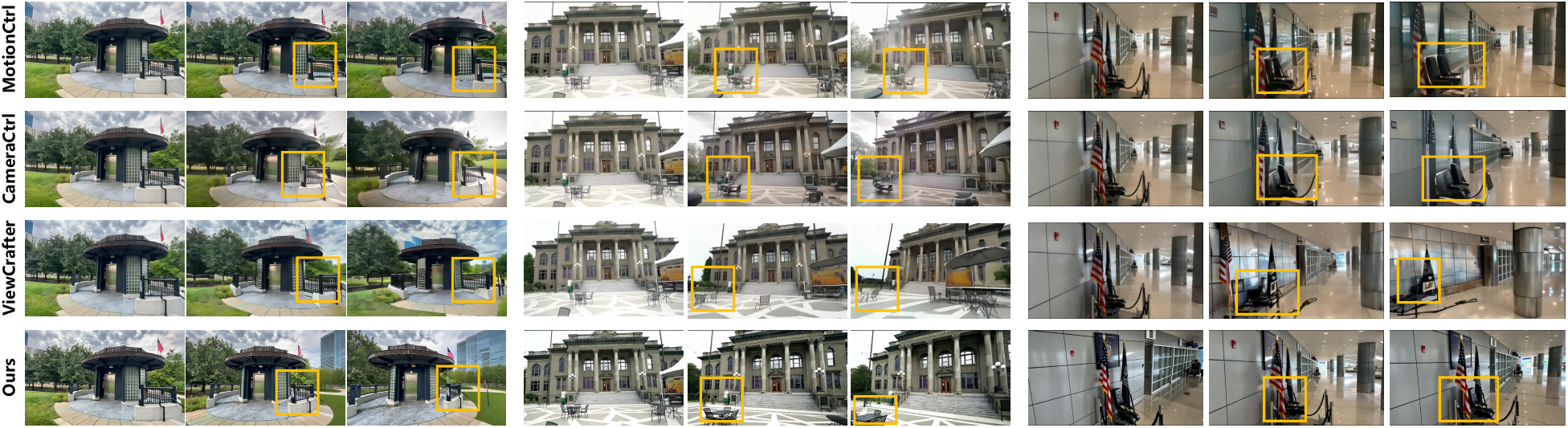}
    \vspace{-2mm}
    \caption{\textbf{Qualitative comparison of generated videos.} Compared to other benchmark methods, the videos generated by our method have better foreground object-background consistency, while also being able to generate videos with larger motion amplitude.}
    \vspace{-2mm}
    \label{fig:video_compare}
\end{figure*}

\subsection{Video-based Scene Reconstruction}
\label{sec:scene_recon}
\noindent{\textbf{Motivation.}}
There are several challenges in reconstructing scenes directly from generated videos. 
First, the limited number of generated frames makes it difficult to capture complete scene information.
In addition, the diverse styles of video frames can pose a challenge to traditional reconstruction techniques.
Furthermore, the generated video frames may contain imperfect or low-quality areas, making the reconstruction process unstable.
Moreover, In stylized scenes, previous methods often yield poor results, as recovering poses and sparse point clouds from video frames is difficult, leading to many artifacts.
To address this, a powerful reconstruction module specifically designed to handle the generated videos is required.

\noindent{\textbf{Latent Feature Fusion.}}
Given input video latents $\mathbf{z} \in \mathbb{R}^{T \times H \times W \times C}$ and camera poses encoded as ray embeddings $\mathbf{p} \in \mathbb{R}^{T \times H \times W \times 6}$, we transform them into token sequences through a patchification process. Specifically, we apply spatial patchification to latent features to obtain latent tokens $\mathbf{z}_t \in \mathbb{R}^{N \times d_z}$, while ray embeddings undergo 3D-patchification to produce pose tokens $\mathbf{p}_t \in \mathbb{R}^{N \times d_p}$.
To incorporate explicit geometric guidance, we leverage monocular depth estimation within our pipeline. For each input RGB video frame, we estimate dense depth maps $\mathbf{D} \in \mathbb{R}^{T \times H \times W \times 1}$, which are processed by a dedicated depth encoder to yield depth tokens $\mathbf{d}_t \in \mathbb{R}^{N \times d_d}$.
The three token sets are channel-wise concatenated to form a unified representation $\mathbf{x} = [\mathbf{z}_t; \mathbf{p}_t; \mathbf{d}_t] \in \mathbb{R}^{N \times (d_z+d_p+d_d)}$, which is then linearly projected to a lower-dimensional space $\mathbf{x}' \in \mathbb{R}^{N \times d}$ before being fed into a sequence of mamba and transformer blocks.
Mamba~\cite{gu2023mamba} achieves the same token sequence processing functionality as Transformer, but reduces computational complexity from $\mathcal{O}(L^2)$ to $\mathcal{O}(L)$, making it particularly suitable for dense reconstruction tasks.
In our Mamba blocks, we implement bi-directional scanning across the token sequence. 
First, we compute state parameters $\mathbf{A}, \mathbf{B}, \mathbf{C} \in \mathbb{R}^{N \times d}$ from the input tokens using a linear projection.
Then, we execute the State Space Model in both forward ($\mathbf{y}_f$) and backward ($\mathbf{y}_b$) directions. The final output of each Mamba block is computed as $\mathbf{y} = \mathbf{y}_f + \mathbf{y}_b$ followed by a final linear transformation.

\noindent{\textbf{Gaussian Decoding.}}
We design a lightweight decoder module that efficiently transforms output feature tokens into per-pixel Gaussian parameters.
The decoding module consists of 3D-DeConv layers, which generates refined Gaussian feature map \( G \in \mathbb{R}^{(T \times H \times W) \times 12} \).
This 12-channel representation precisely encodes the complete set of Gaussian parameters: RGB color (3 channels), scale factors (3 channels), rotation quaternion (4 channels), opacity (1 channel), and ray distance (1 channel). 
The final output of the model is the merge of 3D Gaussians from all input video frames.

\noindent{\textbf{Training Objective.}}
%
%
During training, we render images from predicted Gaussians using randomly selected supervision views. 
Our approach employs a composite loss that integrates three components:
\begin{equation}
    \mathcal{L}_{recon} = \lambda_1\mathcal{L}_{mse} + \lambda_2\mathcal{L}_{perc} + \lambda_3\mathcal{L}_{depth},
\end{equation}
where $\mathcal{L}_{mse}$ represents pixel-wise mean squared error, $\mathcal{L}_{perc}$ denotes perceptual loss, and $\mathcal{L}_{depth}$ enforces depth consistency. 
This formulation jointly optimizes for photometric accuracy, high-level perceptual fidelity, and geometric coherence.

\begin{table}[t!]
\centering
\resizebox{\columnwidth}{!}{%
\begin{tabular}{lccccccc}
\toprule
\textbf{Method} &
  \cellcolor[HTML]{FEF1F1}\textbf{FVD$\downarrow$} &
  \cellcolor[HTML]{FEF1F1}\textbf{FID$\downarrow$} &
  \cellcolor[HTML]{FEF1F1}\textbf{R$_{err} \downarrow$} &
  \cellcolor[HTML]{FEF1F1}\textbf{T$_{err} \downarrow$} &
  \cellcolor[HTML]{FEF1F1}\textbf{LPIPS $\downarrow$} &
  \cellcolor[HTML]{F0FBEF}\textbf{PSNR $\uparrow$} &
  \cellcolor[HTML]{F0FBEF}\textbf{SSIM $\uparrow$} \\ \midrule
\textbf{RealEstate10K} &
  \multicolumn{1}{l}{} &
  \multicolumn{1}{l}{} &
  \multicolumn{1}{l}{} &
  \multicolumn{1}{l}{} &
  \multicolumn{1}{l}{} &
  \multicolumn{1}{l}{} &
  \multicolumn{1}{l}{} \\
MotionCtrl~\cite{wang2023motionctrl} &
  22.65 &
  230.12 &
  0.234 &
  0.798 &
  0.299 &
  14.72 &
  0.404 \\
CameraCtrl~\cite{he2024cameractrl} &
  21.48 &
  188.21 &
  0.054 &
  0.127 &
  0.230 &
  17.33 &
  0.516 \\
ViewCrafter~\cite{yu2024viewcrafter} &
  20.96 &
  204.18 &
  0.055 &
  0.153 &
  0.215 &
  18.95 &
  0.503 \\
Ours &
  \textbf{18.25} &
  \textbf{183.25} &
  \textbf{0.052} &
  \textbf{0.103} &
  \textbf{0.207} &
  \textbf{19.21} &
  \textbf{0.523} \\ \midrule
\textbf{Tanks-and-Temples} &
  \multicolumn{1}{l}{} &
  \multicolumn{1}{l}{} &
  \multicolumn{1}{l}{} &
  \multicolumn{1}{l}{} &
  \multicolumn{1}{l}{} &
  \multicolumn{1}{l}{} &
  \multicolumn{1}{l}{} \\
MotionCtrl~\cite{wang2023motionctrl} &
  30.25 &
  290.38 &
  0.838 &
  1.505 &
  0.315 &
  14.64 &
  0.388 \\
CameraCtrl~\cite{he2024cameractrl} &
  24.41 &
  244.82 &
  0.118 &
  0.294 &
  0.285 &
  15.38 &
  0.469 \\
ViewCrafter~\cite{yu2024viewcrafter} &
  22.50 &
  231.35 &
  0.126 &
  0.307 &
  0.247 &
  \textbf{16.24} &
  \textbf{0.508} \\
Ours &
  \textbf{20.17} &
  \textbf{192.52} &
  \textbf{0.097} &
  \textbf{0.175} &
  \textbf{0.228} &
  16.12 &
  0.501 \\ \midrule
\textbf{DL3DV} &
  \multicolumn{1}{l}{} &
  \multicolumn{1}{l}{} &
  \multicolumn{1}{l}{} &
  \multicolumn{1}{l}{} &
  \multicolumn{1}{l}{} &
  \multicolumn{1}{l}{} &
  \multicolumn{1}{l}{} \\
MotionCtrl~\cite{wang2023motionctrl} &
  25.65 &
  249.51 &
  0.473 &
  1.118 &
  0.312 &
  14.38 &
  0.386 \\
CameraCtrl~\cite{he2024cameractrl} &
  22.76 &
  233.54 &
  0.095 &
  0.238 &
  0.262 &
  16.33 &
  0.489 \\
ViewCrafter~\cite{yu2024viewcrafter} &
  20.59 &
  211.24 &
  0.093 &
  0.244 &
  0.242 &
  \textbf{17.12} &
  0.521 \\
Ours &
  \textbf{18.21} &
  \textbf{171.52} &
 \textbf{0.063} &
  \textbf{0.134} &
  \textbf{0.225} &
  17.02 &
  \textbf{0.537} \\ \bottomrule
\end{tabular}%
}
\vspace{-2mm}
\caption{\textbf{Quantitative comparison} to the camera conditioned video generation method on RealEstate10K~\cite{realestate10k}, DL3DV~\cite{ling2024dl3dv}, and Tanks-and-Temples~\cite{tankandtemple} dataset.}
\vspace{-2mm}

\label{tab:video_quality}
\end{table}

\section{Experiment}
\subsection{Experiment settings}
\noindent{\textbf{Datasets.}}
To comprehensively capture the underlying distribution of real-world scenarios, we trained our video diffusion model on three diverse datasets: RealEstate-10K~\cite{realestate10k}, ACID~\cite{liu2021infiniteacid}, and DL3DV-10K~\cite{ling2024dl3dv}.
The Re10K dataset from YouTube comprises 67,477 training and 7,289 testing indoor and outdoor camera trajectories. 
The ACID dataset focuses on natural landscapes, with 11,075 training and 1,972 testing scenes. 
The DL3DV-10K dataset is a large-scale collection featuring 10,510 scenes captured under controlled, standardized conditions.
\begin{figure*}[t]  
    \centering  
    \begin{minipage}{0.49\textwidth}  
        \centering  
        \includegraphics[width=\textwidth]{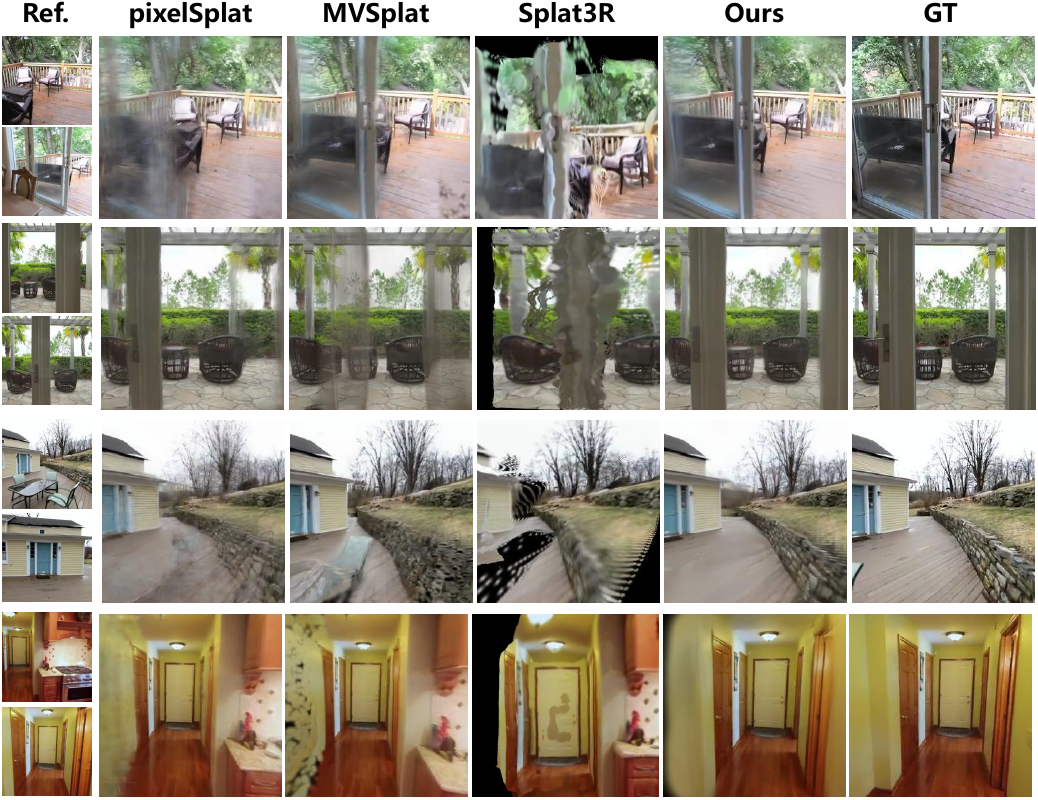}  
        \vspace{-2mm}  
        \caption{\textbf{Qualitative comparison} on Re10K~\cite{realestate10k}. Compared to baselines, we obtain superior reconstruction from limited overlap, and enhanced geometry reconstruction in non-overlapping regions.}  
        \vspace{-4mm}  
        \label{fig:3d_res_1}  
    \end{minipage}  
    \hfill  
    \begin{minipage}{0.49\textwidth}  
        \centering  
        \includegraphics[width=\textwidth]{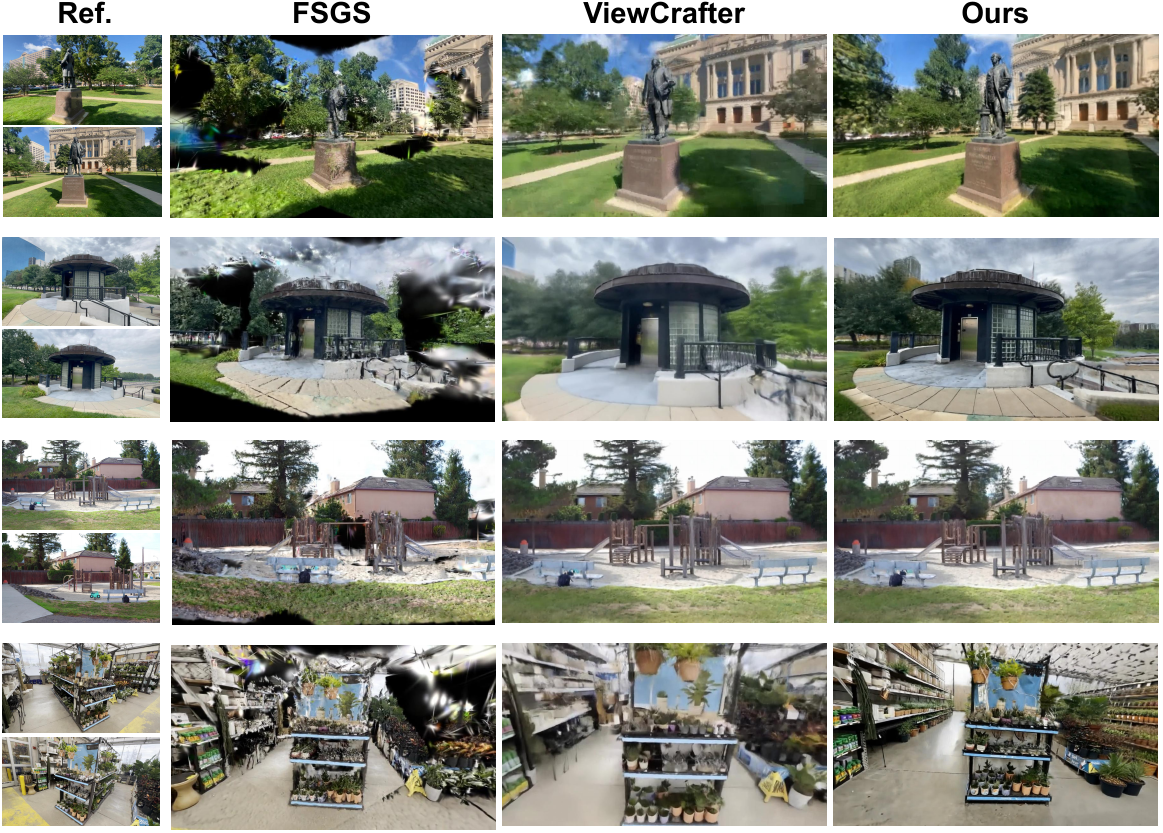}  
        \vspace{-2mm}  
        \caption{\textbf{Qualitative comparison} on DL3DV and Tank-and-Temples dataset. Our method reconstructs better than baselines, even with limited image overlap and in non-overlapping regions.}  
        \vspace{-2mm}  
        \label{fig:3d_res_2}  
    \end{minipage}  
\end{figure*}



\noindent{\textbf{Implementation Details.}}
Our video generation model is based on SVD~\cite{blattmann2023stablesvd}, an image-video diffusion model based on UNet.
In our experiments, we use a relative camera system in which all camera poses are converted to poses relative to the first frame. The camera in the first frame is located at the world origin.
In the first stage, we train the model at a resolution of $320\times 512$ for 50,000 iterations with the frame length set to 25.
Subsequently, we train on $576\times 1024$ for 10,000 iterations to adapt to high resolution.
The learning rate is set to $1e-5$ with a warmup of 1,000 steps, using the Adam optimizer.
We chose Lightning as the training framework, using mixed-precision fp16 and DeepSpeed ZeRO-2.
We trained the proposed method and its variants on 16 NVIDIA A800 GPUs with a batch size of 32.
During inference, we adopt DDIM sampler~\cite{ddim} with classifier-free guidance~\cite{classfree}.

\noindent{\textbf{Progressive Training.}}
Controlling a video generation model to generate arbitrary motion trajectories remains a challenging task.
To enable robust arbitrary trajectory generation, we employ a three-stage curriculum learning strategy. 
The model first trains on smooth camera motions with small temporal intervals, then progressively adapts to more complex motion patterns through linear interval scheduling, and finally incorporates random sampling intervals. 
This gradual transition from simple to complex trajectories proves crucial for high-quality video generation with arbitrary trajectories.

\begin{table}[t!]
\centering
\resizebox{\columnwidth}{!}{%
\begin{tabular}{lccccccccc}
\toprule
 &
  \multicolumn{3}{c}{3-view} &
  \multicolumn{3}{c}{6-view} &
  \multicolumn{3}{c}{9-view} \\ \cmidrule(l){2-10} 
\multirow{-2}{*}{\textbf{Method}} &
  \cellcolor[HTML]{F0FBEF}\textbf{PSNR ↑} &
  \cellcolor[HTML]{F0FBEF}\textbf{SSIM ↑} &
  \cellcolor[HTML]{FEF1F1}\textbf{LPIPS ↓} &
  \cellcolor[HTML]{F0FBEF}\textbf{PSNR ↑} &
  \cellcolor[HTML]{F0FBEF}\textbf{SSIM ↑} &
  \cellcolor[HTML]{FEF1F1}\textbf{LPIPS ↓} &
  \cellcolor[HTML]{F0FBEF}\textbf{PSNR ↑} &
  \cellcolor[HTML]{F0FBEF}\textbf{SSIM ↑} &
  \cellcolor[HTML]{FEF1F1}\textbf{LPIPS ↓} \\ \midrule
\textbf{Mip-NeRF360} &
  \multicolumn{1}{l}{} &
  \multicolumn{1}{l}{} &
  \multicolumn{1}{l}{} &
  \multicolumn{1}{l}{} &
  \multicolumn{1}{l}{} &
  \multicolumn{1}{l}{} &
  \multicolumn{1}{l}{} &
  \multicolumn{1}{l}{} &
  \multicolumn{1}{l}{} \\
Zip-NeRF~\cite{barron2023zipnerf} &
  12.77 &
  0.271 &
  0.705 &
  13.61 &
  0.284 &
  0.663 &
  14.30 &
  0.312 &
  0.633 \\
ReconFusion~\cite{wu2024reconfusion} &
  15.50 &
  0.358 &
  0.585 &
  16.93 &
  0.401 &
  0.544 &
  18.19 &
  0.432 &
  0.511 \\
CAT3D~\cite{gao2024cat3d} &
  16.62 &
  0.377 &
  0.515 &
  17.72 &
  0.425 &
  0.482 &
  18.67 &
  0.460 &
  0.460 \\
ReconX~\cite{liu2024reconx} &
  17.16 &
  0.435 &
  0.407 &
  19.20 &
  0.473 &
  0.378 &
  20.13 &
  0.482 &
  0.356 \\
ViewCrafter~\cite{yu2024viewcrafter} &
  14.51 &
  0.315 &
  0.682 &
  15.87 &
  0.336 &
  0.665 &
  16.45 &
  0.348 &
  0.655 \\
Ours &
  \textbf{17.32} &
  \textbf{0.439} &
  \textbf{0.418} &
  \textbf{19.42} &
  \textbf{0.510} &
  \textbf{0.371} &
  \textbf{20.19} &
  \textbf{0.536} &
  \textbf{0.345} \\ \midrule
\textbf{DTU} &
  \multicolumn{1}{l}{} &
  \multicolumn{1}{l}{} &
  \multicolumn{1}{l}{} &
  \multicolumn{1}{l}{} &
  \multicolumn{1}{l}{} &
  \multicolumn{1}{l}{} &
  \multicolumn{1}{l}{} &
  \multicolumn{1}{l}{} &
  \multicolumn{1}{l}{} \\
Zip-NeRF~\cite{barron2023zipnerf} &
  9.18 &
  0.601 &
  0.383 &
  8.84 &
  0.589 &
  0.370 &
  9.23 &
  0.592 &
  0.364 \\
FSGS~\cite{zhu2023fsgs} &
  17.34 &
  0.818 &
  0.169 &
  21.55 &
  0.880 &
  0.127 &
  24.33 &
  0.911 &
  0.106 \\
ReconFusion~\cite{wu2024reconfusion} &
  20.74 &
  0.875 &
  0.124 &
  23.62 &
  \textbf{0.904} &
  0.105 &
  24.62 &
  0.921 &
  0.094 \\
CAT3D~\cite{gao2024cat3d} &
  22.02 &
  0.844 &
  0.121 &
  24.28 &
  0.899 &
  0.095 &
  25.92 &
  \textbf{0.928} &
  0.073 \\
ViewCrafter~\cite{yu2024viewcrafter} &
  15.63 &
  0.522 &
  0.383 &
  15.03 &
  0.525 &
  0.408 &
  14.92 &
  0.487 &
  0.452 \\
Ours &
  \textbf{27.92} &
  \textbf{0.879} &
  \textbf{0.103} &
  \textbf{28.82} &
  0.895 &
  \textbf{0.082} &
  \textbf{29.03} &
  0.905 &
  \textbf{0.065} \\ \midrule
\textbf{LLFF} &
  \multicolumn{1}{l}{} &
  \multicolumn{1}{l}{} &
  \multicolumn{1}{l}{} &
  \multicolumn{1}{l}{} &
  \multicolumn{1}{l}{} &
  \multicolumn{1}{l}{} &
  \multicolumn{1}{l}{} &
  \multicolumn{1}{l}{} &
  \multicolumn{1}{l}{} \\
Zip-NeRF~\cite{barron2023zipnerf} &
  17.23 &
  0.574 &
  0.373 &
  20.71 &
  0.764 &
  0.221 &
  23.63 &
  0.830 &
  0.166 \\
FSGS~\cite{zhu2023fsgs} &
  20.31 &
  0.652 &
  0.288 &
  24.20 &
  0.811 &
  0.173 &
  25.32 &
  0.856 &
  0.136 \\
ReconFusion~\cite{wu2024reconfusion} &
  21.34 &
  0.724 &
  0.203 &
  24.25 &
  0.815 &
  0.152 &
  25.21 &
  0.848 &
  0.134 \\
CAT3D~\cite{gao2024cat3d} &
  21.58 &
  0.731 &
  0.181 &
  24.71 &
  \textbf{0.833} &
  \textbf{0.121} &
  25.63 &
  \textbf{0.860} &
  0.107 \\
ViewCrafter~\cite{yu2024viewcrafter} &
  17.73 &
  0.521 &
  0.332 &
  17.43 &
  0.512 &
  0.345 &
  17.33 &
  0.488 &
  0.351 \\
Ours &
  \textbf{22.04} &
  \textbf{0.741} &
  \textbf{0.165} &
  \textbf{25.11} &
  0.814 &
  0.124 &
  \textbf{25.95} &
  0.838 &
  \textbf{0.103} \\ \bottomrule
\end{tabular}%
}
\vspace{-2mm}
\caption{\textbf{Quantitative comparison} of sparse view 3D reconstruction on Out-of-Domain Datasets.}
\vspace{-6mm}
\label{tab:sparse_view_recon}
\end{table}

\subsection{Comparisons}

\subsubsection{Controllable Video Generation}
\noindent{\textbf{Benchmark and Metrics.}}
We evaluate our video generation model against three baselines~\cite{wang2023motionctrl,he2024cameractrl,yu2024viewcrafter} on three benchmark datasets: RealEstate10K~\cite{realestate10k} (500 randomly selected videos with first frame as image condition and subsequent frames at stride 3 for pose guidance), DL3DV-140~\cite{ling2024dl3dv} (500 video clips at stride 2), and Tanks-and-Temples~\cite{tankandtemple} (100 sequences at stride 4 from 14 scenes with COLMAP-annotated poses) for out-of-domain generalization testing.

\noindent{\textbf{Metrics.}}
Our evaluation framework employs multiple metrics to assess performance: visual quality measured through Fréchet Video Distance (FVD) and Fréchet Inception Distance (FID); camera control precision quantified by rotation error ($R_{err}$) and translation error ($T_{err}$) computed from camera poses extracted via COLMAP and normalized relative to the first frame; and visual consistency evaluated using PSNR, SSIM, and LPIPS~\cite{lpips} between generated and ground-truth views. To ensure fair comparison across methods with varying output capabilities, we restrict visual similarity assessment to the first 14 frames, as generated content typically diverges progressively from the conditional single-view input as the scene extends.


\begin{table}[t!]
\centering
\resizebox{\columnwidth}{!}{%
\begin{tabular}{l|ccc|ccc}
\toprule
            & \multicolumn{3}{c|}{\textbf{RealEstate10K}}      & \multicolumn{3}{c}{\textbf{ACID}}                \\ \cmidrule(l){2-7} 
\multirow{-2}{*}{\textbf{Method}} &
  \cellcolor[HTML]{F0FBEF}\textbf{PSNR ↑} &
  \cellcolor[HTML]{F0FBEF}\textbf{SSIM ↑} &
  \cellcolor[HTML]{FEF1F1}\textbf{LPIPS ↓} &
  \cellcolor[HTML]{F0FBEF}\textbf{PSNR ↑} &
  \cellcolor[HTML]{F0FBEF}\textbf{SSIM ↑} &
  \cellcolor[HTML]{FEF1F1}\textbf{LPIPS ↓} \\ \midrule
pixelNeRF~\cite{yu2021pixelnerf}   & 20.43          & 0.589          & 0.550          & 20.97          & 0.547          & 0.533          \\
GPNR~\cite{suhail2022generalizablegpnr}        & 24.11          & 0.793          & 0.255          & 25.28          & 0.764          & 0.332          \\
AttnRend~\cite{attnrend}    & 24.78          & 0.820          & 0.213          & 26.88          & 0.799          & 0.218          \\
MuRF~\cite{xu2024murf}        & 26.10          & 0.858          & 0.143          & 28.09          & 0.841          & 0.155          \\
pixelSplat~\cite{charatan2024pixelsplat}  & 25.89          & 0.858          & 0.142          & 28.14          & 0.839          & 0.150          \\
MVSplat~\cite{chen2024mvsplat}     & 26.39          & 0.869          & 0.128          & 28.25          & 0.843          & 0.144          \\
GS-LRM~\cite{gslrm}      & 28.10          & 0.892          & 0.114          & -              & -              & -              \\
DepthSplat~\cite{xu2024depthsplat}  & 24.23          & 0.790          & 0.217          & -              & -              & -              \\
ReconX~\cite{liu2024reconx}      & 28.31          & \textbf{0.912} & \textbf{0.088} & 28.84          & 0.891          & \textbf{0.101} \\
ViewCrafter~\cite{yu2024viewcrafter} & 24.22          & 0.788          & 0.218          & 23.48          & 0.660          & 0.299          \\
Ours        & \textbf{28.35} & 0.862          & 0.121          & \textbf{28.93} & \textbf{0.899} & 0.116          \\ \bottomrule
\end{tabular}%
}
\vspace{-2mm}
\caption{\textbf{Quantitative comparison} of two-view sparse view reconstruction methods on RealEstate10K~\cite{realestate10k} and ACID~\cite{liu2021infiniteacid} dataset.}
\vspace{-2mm}
\label{tab:two_view_sparse_view_recon}
\end{table}

\noindent{\textbf{Comparison.}}
Table~\ref{tab:video_quality} shows that our method excels in qualitative results, outperforming other methods in terms of visual quality, pose control, and 3D consistency. 
Figure~\ref{fig:video_compare} further demonstrates our advantage in 3D consistency and visual quality. 
This performance improvement is attributed to the introduction of a camera parameterization method and an epipolar attention module, which respectively enhance pose control ability and the 3D consistency of generated videos. 
In addition, training on a diverse scene dataset significantly improves the generalization ability of the model.

\begin{table}[t!]
\centering
\resizebox{\columnwidth}{!}{%
\begin{tabular}{@{}l|ccc|ccc@{}}
\toprule
              & \multicolumn{3}{c|}{Small}                       & \multicolumn{3}{c}{Large}                         \\ \cmidrule(l){2-7} 
\multirow{-2}{*}{Method} &
  \cellcolor[HTML]{F0FBEF}PSNR$\uparrow$ &
  \cellcolor[HTML]{F0FBEF}SSIM$\uparrow$ &
  \cellcolor[HTML]{FEF1F1}LPIPS$\downarrow$ &
  \cellcolor[HTML]{F0FBEF}PSNR$\uparrow$ &
  \cellcolor[HTML]{F0FBEF}SSIM$\uparrow$ &
  \cellcolor[HTML]{FEF1F1}LPIPS$\downarrow$ \\ \midrule
pixelNeRF~\cite{yu2021pixelnerf}     & 18.417          & 0.601          & 0.526         & 20.869          & 0.639          & 0.458          \\
AttnNeRF~\cite{attnrend}      & 19.151          & 0.663          & 0.368         & 25.897          & 0.845          & 0.229          \\
pixelSplat~\cite{charatan2024pixelsplat}    & 20.263          & 0.717          & 0.266         & 27.151          & 0.879          & 0.122          \\
MVSplat~\cite{chen2024mvsplat}       & 20.353          & 0.724          & 0.254          & 27.408          & 0.884          & 0.116          \\
DUSt3R~\cite{wang2024dust3r}        & 14.101          & 0.432          & 0.468         & 16.427          & 0.453          & 0.402          \\
CoPoNeRF~\cite{hong2024coponerf}      & 17.393          & 0.585          & 0.462         & 20.464          & 0.652          & 0.358          \\
\textbf{Ours} & \textbf{22.514} & \textbf{0.784} & \textbf{0.213} & \textbf{27.411} & \textbf{0.913} & \textbf{0.109} \\ \bottomrule
\end{tabular}%
}
\vspace{-2mm}
\caption{\textbf{Quantitative comparison} on the RealEstate10K~\cite{realestate10k} dataset. \textit{Small} and \textit{Large} refer to input images with low and high overlap ratios, respectively. Greater overlap means greater temporal coherence between adjacent images.}
\vspace{-2mm}
\label{tab:sparse_recon_re10k}
\end{table}
\begin{table}[t!]
\centering
\resizebox{\columnwidth}{!}{%
\begin{tabular}{l|ccc|ccc}
\toprule
              & \multicolumn{3}{c|}{Tanks-and-Temples}            & \multicolumn{3}{c}{DL3DV}                         \\ \cmidrule(l){2-7} 
\multirow{-2}{*}{Dataset} &
  \cellcolor[HTML]{F0FBEF}PSNR$\uparrow$ &
  \cellcolor[HTML]{F0FBEF}SSIM$\uparrow$ &
  \cellcolor[HTML]{FEF1F1}LPIPS$\downarrow$ &
  \cellcolor[HTML]{F0FBEF}PSNR$\uparrow$ &
  \cellcolor[HTML]{F0FBEF}SSIM$\uparrow$ &
  \cellcolor[HTML]{FEF1F1}LPIPS$\downarrow$ \\ \midrule
LucidDreamer~\cite{chung2023luciddreamer}  & 14.552          & 0.364          & 0.415          & 15.126          & 0.452          & 0.431          \\
ZeroNVS~\cite{zeronvs}       & 14.734          & 0.381          & 0.480          & 15.163          & 0.464          & 0.466          \\
MotionCtrl~\cite{wang2023motionctrl}    & 15.322          & 0.426          & 0.404          & 16.889          & 0.528          & 0.392          \\
ViewCrafter~\cite{yu2024viewcrafter}   & 21.286          & 0.654          & 0.193          & 21.373          & 0.686          & 0.244          \\
\textbf{Ours} & \textbf{22.331} & \textbf{0.742} & \textbf{0.213} & \textbf{22.782} & \textbf{0.791} & \textbf{0.202} \\ \bottomrule
\end{tabular}%
}
\vspace{-2mm}
\caption{\textbf{Quantitative comparison} of single-view novel view synthesis on DL3DV~\cite{ling2024dl3dv} and Tank-and-Temples~\cite{tankandtemple} benchmarks.}
\vspace{-4mm}

\label{tab:single_3d}
\end{table}

\subsubsection{Sparse View Reconstruction}
\noindent{\textbf{Benchmark and Metrics.}}
%
To comprehensively evaluate the performance of our method, we compare it with two categories of reconstruction methods: (1) two-view feed-forward reconstruction methods~\cite{charatan2024pixelsplat,attnrend,chen2024mvsplat,gslrm,xu2024depthsplat,suhail2022generalizablegpnr}, and (2) optimization-based sparse-view reconstruction methods~\cite{fan2024instantsplat,gao2024cat3d,yu2024viewcrafter,liu2024reconx}.
The evaluation employs two types of test sets: in-domain test sets (from RealEstate10K~\cite{realestate10k} and Acid~\cite{liu2021infiniteacid}) and out-of-domain test sets. 
To ensure a fair comparison, the out-of-domain test set is the same as that used in previous work~\cite{wu2024reconfusion,gao2024cat3d}.
For evaluating scene reconstruction quality, we adopt three metrics: PSNR, SSIM, and LPIPS~\cite{lpips}. 
Additionally, we categorize the evaluation into two groups based on the overlap between input views, with detailed overlap estimation methods provided in the supplementary materials.
The evaluation protocol follows a consistent procedure: first, we reconstruct the 3D scene based on the input images, then render from novel views, and compute the metrics between the rendered images and the reference images.

\noindent{\textbf{Comparison.}}
%
As shown in Tables~\ref{tab:sparse_view_recon}-\ref{tab:sparse_recon_re10k} and Figures~\ref{fig:3d_res_1}-\ref{fig:3d_res_2}, our method performs well on various evaluation metrics, particularly in challenging scenarios with occlusions and limited view overlap. 
This improvement can be attributed to our approach that utilizes scene priors and sparse input views to generate helpful additional observations, thereby enhancing visual correlations between different viewpoints and helping to reduce the complexity of sparse-view reconstruction.
Compared to the method~\cite{yu2024viewcrafter} conditioning on point-based rendering, our approach shows advantages in reconstructing scenes with thin structures, leading to improved 3D reconstruction results.

\begin{figure}[t!]
    \centering
    \includegraphics[width=0.48\textwidth]{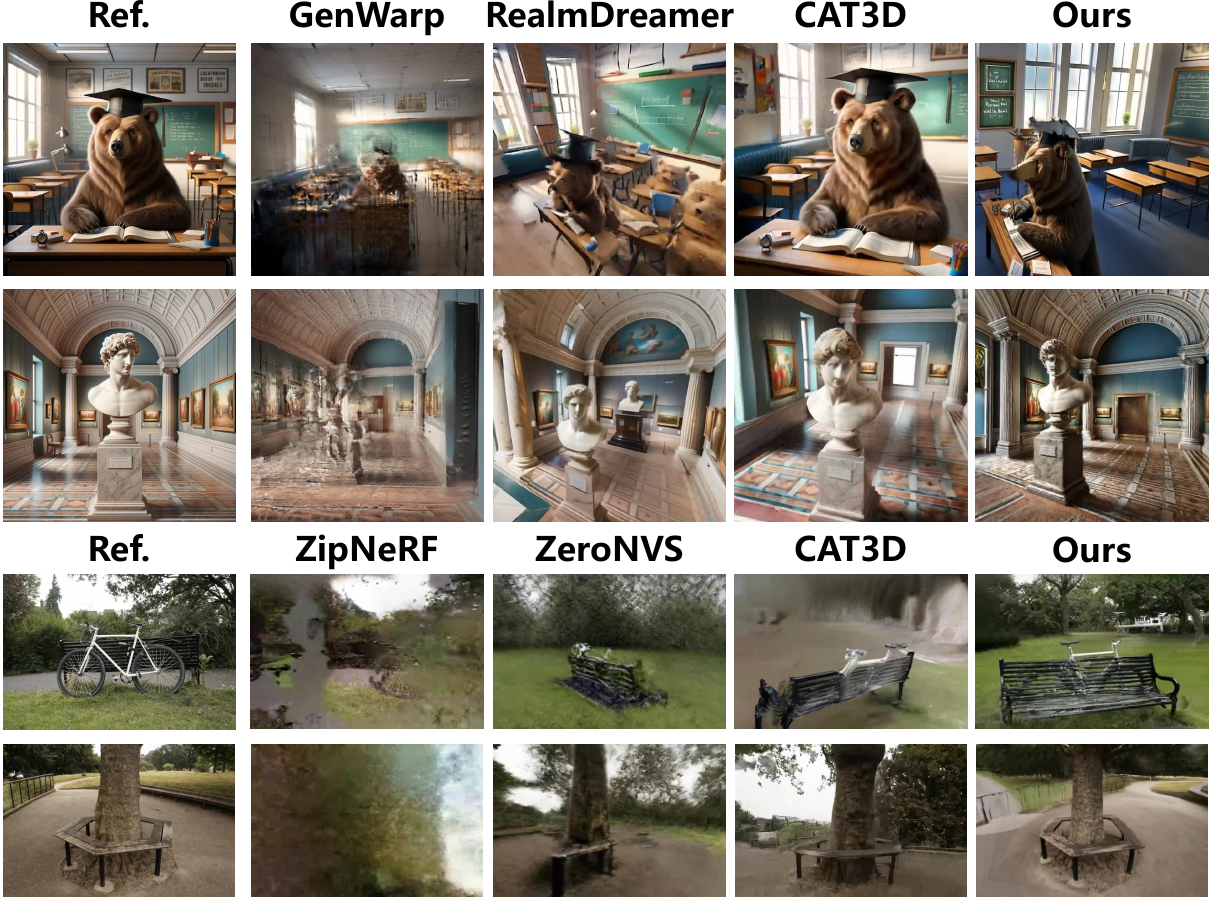}
    \vspace{-2mm}
    \caption{\textbf{Qualitative comparisons} of novel views rendered from scenes reconstructed using other 3D generation methods.}
    \vspace{-4mm}
    \label{fig:single_view_3d}
\end{figure}

\begin{figure*}[t]
    \centering
    \includegraphics[width=1.0\textwidth]{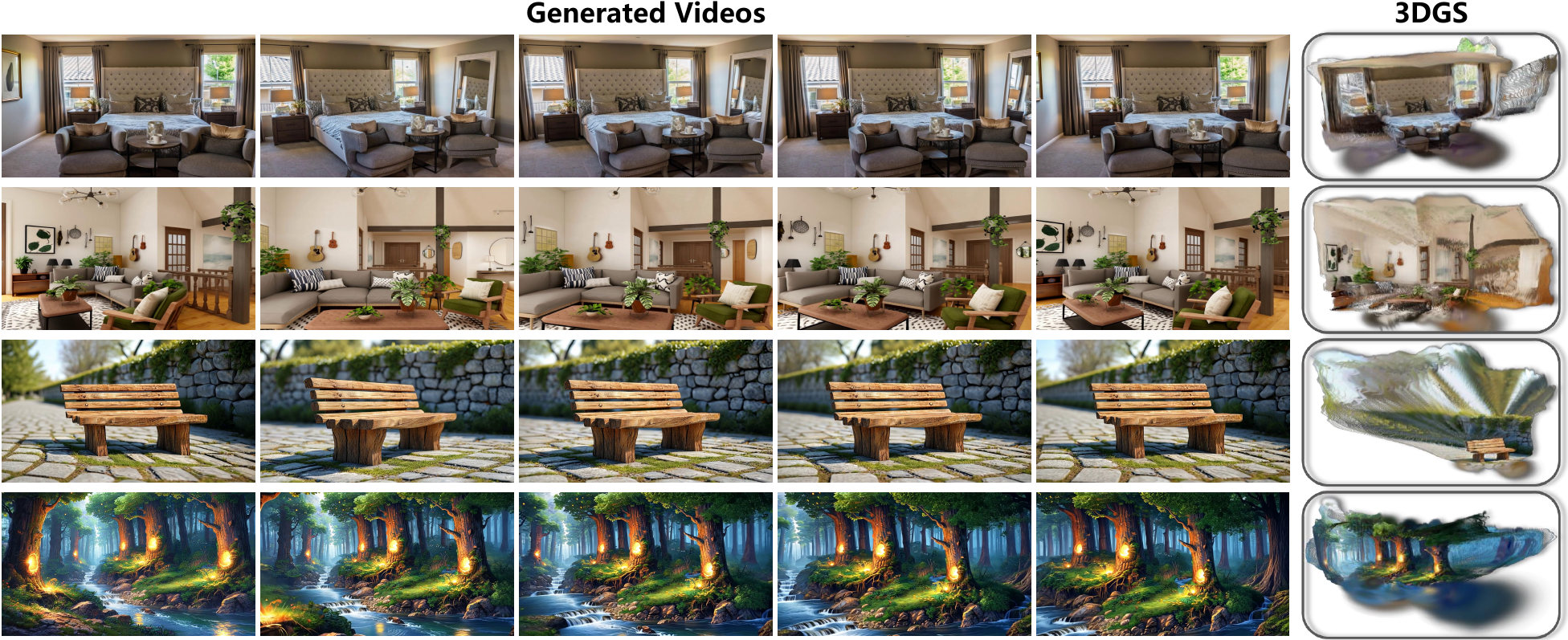}
    \vspace{-2mm}
    \caption{Additional visualizations of generated videos and their corresponding 3DGS representations.}
    \vspace{-2mm}
    \label{fig:video_recon}
\end{figure*}
\subsubsection{Single View 3D Generation}
\noindent{\textbf{Benchmark and Metrics.}}
In single-view generation, we compare our method with several generative methods~\cite{zeronvs,wang2023motionctrl,yu2024viewcrafter,seo2024genwarp} on the Tank-and-Temples~\cite{tankandtemple} and DL3DV-140~\cite{ling2024dl3dv} datasets.
Noted that we begin by reconstructing the 3D scene from the given images, followed by calculating the metrics using renderings from novel views.
Similarly, we used the PSNR, SSIM, and LPIPS~\cite{lpips} metrics to evaluate our results.
Evaluation in this underconstrained setting is challenging, as multiple 3D scenes can produce consistent generations for a given view. 
Thus, we measure metrics using only temporally adjacent frames to the conditional image, 14 sampled frames and poses captured after the conditional image.

\noindent{\textbf{Comparison.}}
As shown in Table~\ref{tab:single_3d}, our method consistently achieves superior performance on image quality metrics compared to the baselines.
For a fair comparison with ZeroNVS~\cite{zeronvs} and LucidDreamer~\cite{chung2023luciddreamer}, which are limited to square image inputs, we crop the generated novel views before computing the quantitative metrics.
Figure~\ref{fig:single_view_3d} shows that scenes reconstructed using our method have less noise and distortion in occluded areas.
For more visualizations, see Figure~\ref{fig:video_recon}.




\subsection{Ablation Study}
\noindent{\textbf{Ablation on Video Diffusion Model.}}
We evaluate the effectiveness of the design choices.
As shown in Table~\ref{tab:model_ablation}, ray embedding achieves significant improvements in camera control accuracy and rendering quality compared to other camera encoding methods by introducing denser position encoding and optimizing geometric correspondences. 
Specifically, cross-frame consistency shows notable enhancement.
Furthermore, when incorporating geometric constraints through Epipolar Attention, the system achieves optimal performance in terms of camera controllability and 3D consistency. 


\begin{table}[t]
\centering
\resizebox{\columnwidth}{!}{%
\begin{tabular}{l|cccc}
\toprule
Method & \cellcolor[HTML]{FEF1F1}\textbf{T}$_{err}\downarrow$ & \cellcolor[HTML]{FEF1F1}\textbf{R}$_{err}\downarrow$ & \cellcolor[HTML]{FEF1F1}COLMAP$_{err}\downarrow$ & \cellcolor[HTML]{FEF1F1}FVD$\downarrow$ \\ \midrule
Raw Value Emb.  & 1.592          & 2.376          & 13.5\%         & 81.854          \\
Quaternion Emb. & 1.457          & 2.394          & 13.1\%         & 80.368          \\
W/o Ray Emb.    & 2.683          & 3.145          & 15.7\%         & 112.843         \\
W/o Epipolar    & 1.125          & 2.235          & 12.2\%         & 79.547          \\
W/o Scale Align.      & 1.712 & 2.498 & 14.3\% & 81.679 \\
\textbf{Full model}      & \textbf{1.014} & \textbf{2.218} & \textbf{4.2\%} & \textbf{78.132} \\

\bottomrule
\end{tabular}%
}
\vspace{-2mm}
\caption{\textbf{Ablation study} on the video diffusion model variants.}
\vspace{-2mm}

\label{tab:model_ablation}
\end{table}
\begin{figure}[t!]
    \centering
    \includegraphics[width=0.49\textwidth]{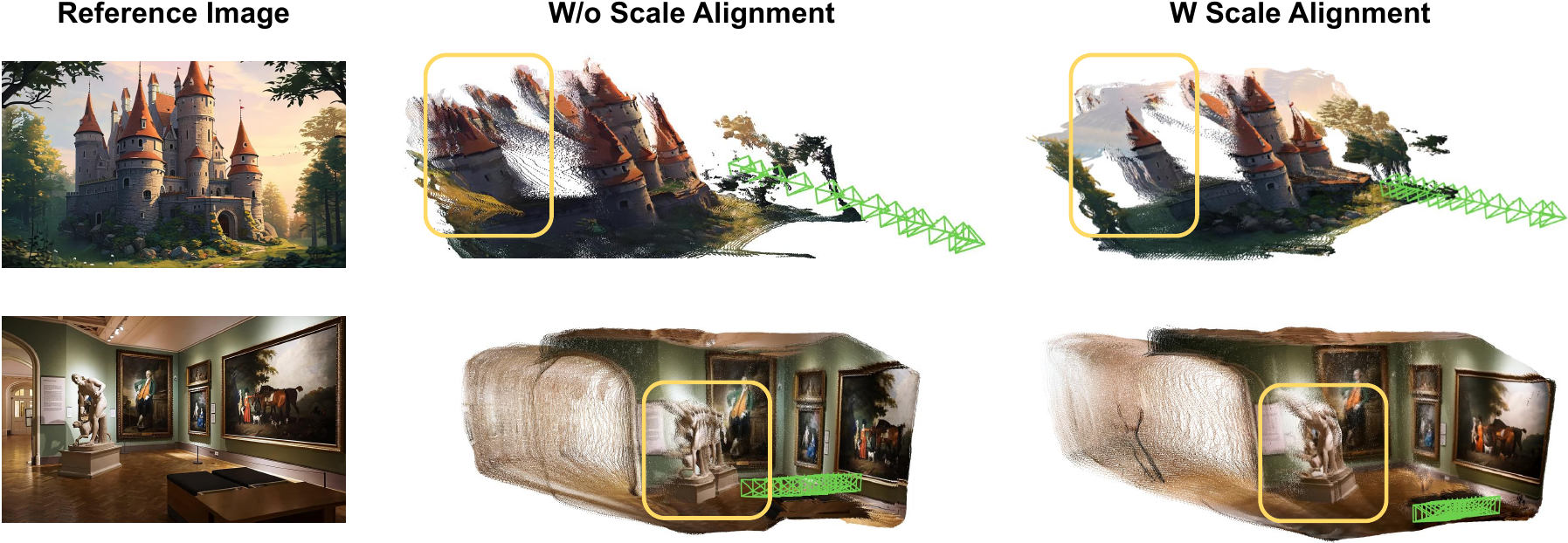}
    \vspace{-2mm}
     \caption{\textbf{Ablation on the Scale Alignment.} The video reconstruction results obtained based on scale-alignment training exhibit reduced artifacts while improving the 3D consistency of the scene.}
    \vspace{-2mm}
    \label{fig:video_3d}
\end{figure}


\noindent{\textbf{Scale Alignment.}}
As illustrated in Table~\ref{tab:model_ablation} and Figure~\ref{fig:video_3d}, the incorporation of dataset metric scale alignment significantly improves the pose control accuracy and 3D consistency of the generated video.

\begin{figure}[t]
    \centering
    \includegraphics[width=0.46\textwidth]{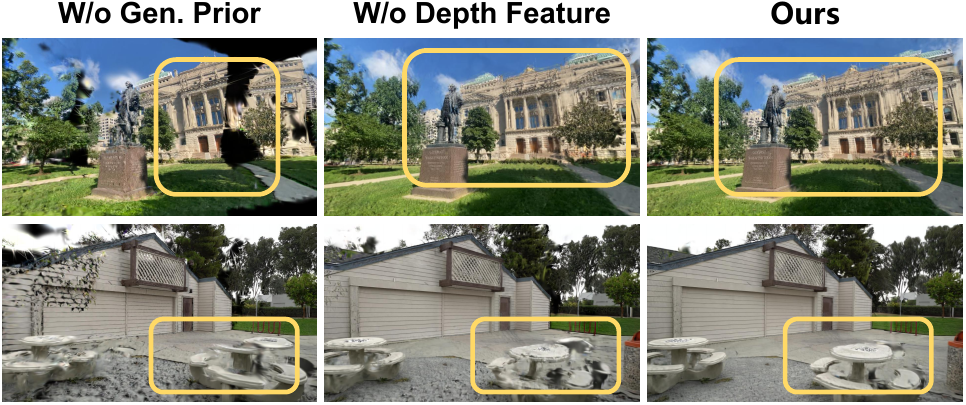}
    \vspace{-2mm}
    \caption{Ablation on the feed-forward Reconstruction Module.}
    \vspace{-2mm}
    \label{fig:geo_ablation}
\end{figure}

\noindent{\textbf{Video Latent-based Reconstruction.}}
As demonstrated in Figure~\ref{fig:geo_ablation},  incorporating generative priors and depth features in our method significantly enhances the fidelity of fine-grained geometric details in scene reconstruction. By introducing priors from video generative models, we reduced the difficulty of sparse-view reconstruction and achieved reasonable completion of unseen regions.



\section{Conclusion}
We present \textit{SpatialCrafter}, a framework for scene reconstruction and novel view synthesis from sparse or single-view inputs.
By leveraging video diffusion models to generate plausible additional observations, we effectively reduce the complexity of sparse-view scene reconstruction.
Our key contributions include: (1) precise camera control via ray embeddings or depth-warped images with a trainable condition encoder; (2) a unified scale estimator solving the scale ambiguity in multi-dataset training; and (3) a hybrid Mamba-Transformer architecture that combines monocular depth priors with semantic features from video latent space to directly regress 3D Gaussian primitives.
Experiments show our method outperforms existing methods, especially in single-view extrapolation and scenarios with little overlaps.
Future work will focus on extending to dynamic scene.


{
    \small
    \bibliographystyle{ieeenat_fullname}
    \bibliography{main}
}


\end{document}